# Reduction of Maximum Entropy Models to Hidden Markov Models

**Joshua Goodman**
Microsoft Research
One Microsoft Way
Redmond, WA 98052
*joshuago@microsoft.com*

## Abstract

We show that maximum entropy (maxent) models can be modeled with certain kinds of HMMs, allowing us to construct maxent models with hidden variables, hidden state sequences, or other characteristics. The models can be trained using the forward-backward algorithm. While the results are primarily of theoretical interest, unifying apparently unrelated concepts, we also give experimental results for a maxent model with a hidden variable on a word disambiguation task; the model outperforms standard techniques.[1]

## 1 Introduction

Maximum Entropy (maxent) models are an attractive formalism for statistical models of many types and have been used for a number of purposes, including language modeling (Rosenfeld, 1994), prepositional phrase attachment (Ratnaparkhi, 1998), sentence breaking (Reynar and Ratnaparkhi, 1997), and parsing (Ratnaparkhi, 1997). Maxent models allow the combination of many different types of information in a principled fashion. They are also called maximum likelihood exponential models, or log-linear models. They can be joint models of the form $P(x)$, or conditional models of the form $P(x|h)$, where $h$ is any set of conditioning variables. We only consider the conditional form, which is more broadly useful. Maxent models are of the form

$$P(x|h) = \frac{\prod_{i=1}^{g} \lambda_i^{f_i(x,h)}}{\sum_y \prod_{i=1}^{g} \lambda_i^{f_i(y,h)}} \qquad (1)$$

where $\lambda_i$ is a weight and $f_i(x, h)$ is an indicator function. We restrict ourselves here to the special, common, case where the $f_i$ take only the values 0 or 1, to indicate whether some condition is true of $x$ and $h$ or not, although the technique could be extended to any integer (but not continuous) valued $f_i$. When trained, maxent models have three important properties (Della Pietra, Della Pietra, and Lafferty, 1997). First, the likelihood of the training data is maximized. Second, on the training data, the expectation of the frequency of each indicator function $f_i$ equals the observation of the frequency:

$$\sum_k \sum_x P(x|h_k) f_i(x, h_k) = \sum_k f_i(x_k, h_k) \qquad (2)$$

Third, the model is as similar as possible to the uniform distribution (minimizes the Kullback-Leibler divergence), given the second constraint, which is why these models are called maximum entropy models.

In Section 2, we show a surprising result: that maxent models can be seen as a special kind of Hidden Markov Model (HMM), with tying of the transition probabilities. There will not, however be a single HMM equivalent to a given maxent model; instead, we will need to build an HMM on the fly for a given history $h$, but the transition probabilities of this HMM will be essentially the same as the weights of the maxent model. We can then use the forward-backward algorithm to train this model, leading to a new training algorithm for maxent models. The result is significant from a theoretical standpoint, because it shows that an important model type, maxent models, can be easily phrased in the HMM formalism, while it would be difficult to efficiently phrase it in the formalisms of graphical models.

When maxent models are phrased as HMMs, it makes it easy to extend and combine the formalisms. In Section 3, we show that we can extend maxent models in a number of ways, adding hidden variables or even hidden state sequences, modeling mixtures of random continuous variables, and even recent formalisms such as Maximum Entropy Markov Models and Conditional Random Fields. While it would have been possible to develop each of these model types individually, by reducing them all to a common, well understood framework, we provide a new theoretical tool for understanding their relationships. We then go on to show experimental results using one of these extensions. Finally, we conclude with a discussion of the implications of this work, especially for future research.

[1] This is the short version of the paper. For the long version, containing complete proofs, more examples, and details of experiments, see `http://www.research.microsoft.com/~joshuago`

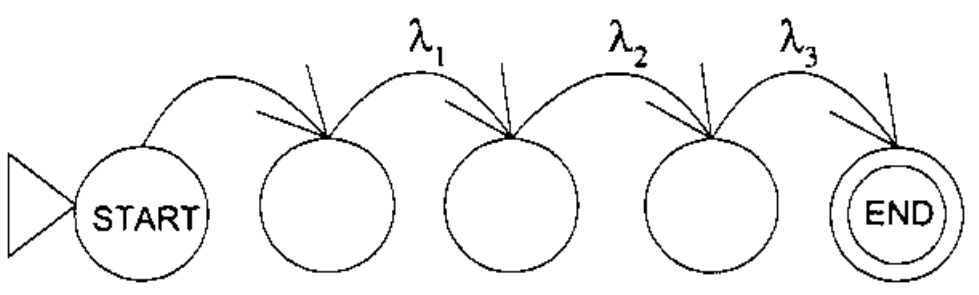

Figure 1: Outputs Proportional to $\lambda_1 \times \lambda_2 \times \lambda_3$.

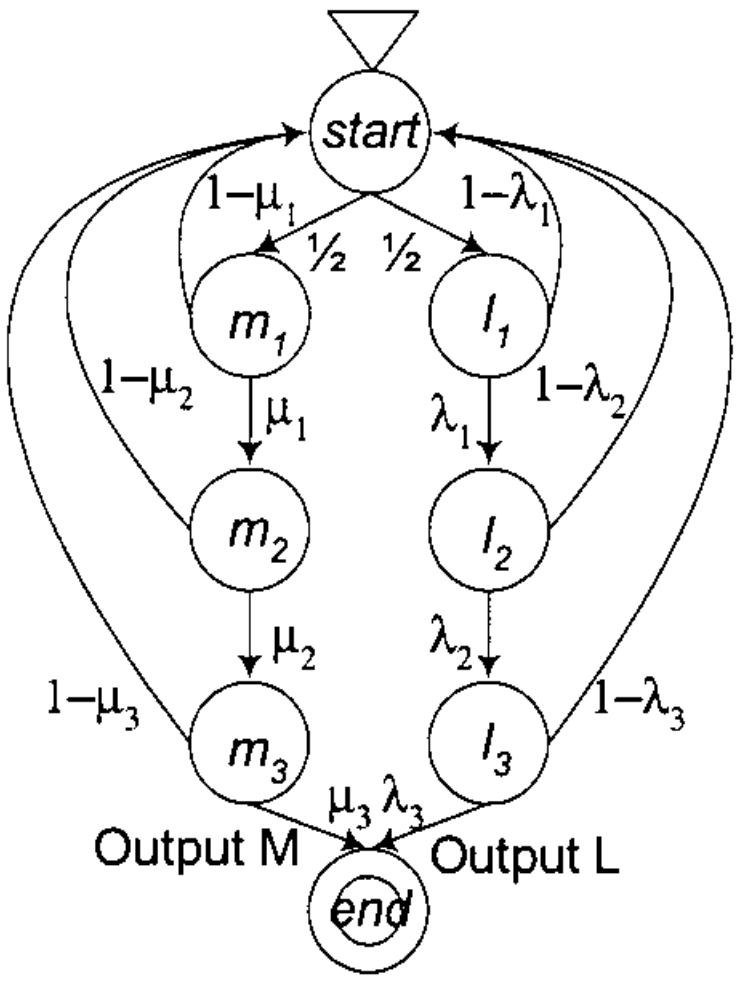

Figure 2: Simple maximum entropy HMM

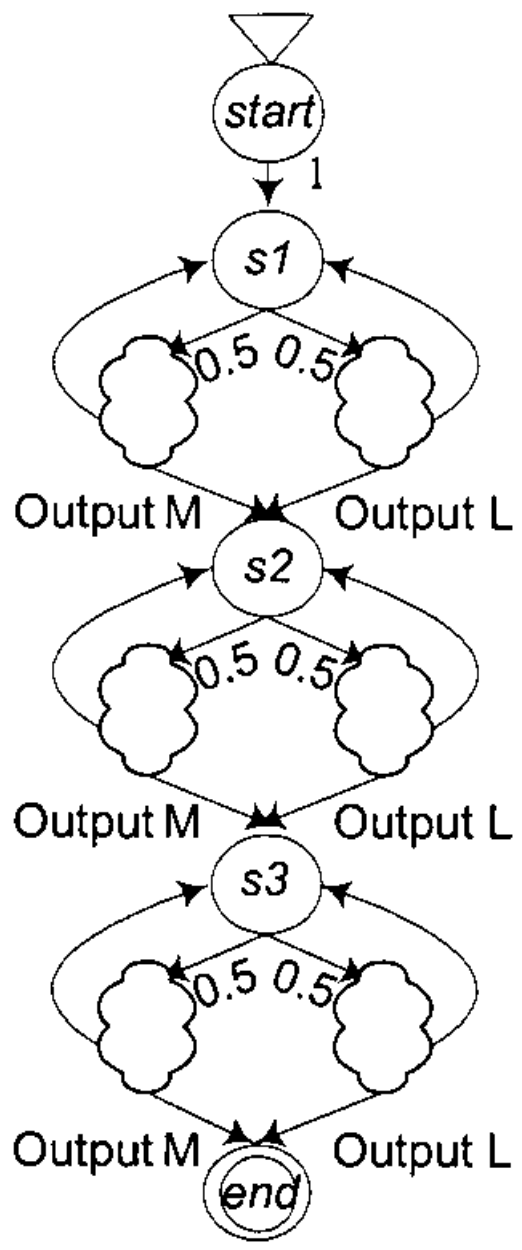

Figure 3: Multiple segments for training

## 2 Reduction to HMMs

In this section, we show the novel result that a maxent model can be phrased as a kind of Hidden Markov Model (HMM) with many tied parameters. Note, however, that for a given maxent model, there is no fixed HMM topology that one uses to find probabilities. Instead, there is a simple algorithm that, for a given maxent model and piece of test data, constructs an HMM that gives the probability for that test data. More importantly, for a given set of training data, we can build an HMM that when trained will converge to the correct parameters.

Throughout this paper, we will use as an example the case where there are two possible values for $x$, L and M (we could have used more typical values, e.g. 0 and 1, but chose L and M to be parallel to the variables $\lambda$ and $\mu$ described below.) We will assume that for L, $h$, there are three indicator functions that are true, with coefficients $\lambda_1, \lambda_2, \lambda_3$, and, similarly, for M, $h$ the coefficients of the true indicators are $\mu_1, \mu_2, \mu_3$. Substituting these values into Equation 1, we get $P(L|h) = \frac{\lambda_1\lambda_2\lambda_3}{\lambda_1\lambda_2\lambda_3+\mu_1\mu_2\mu_3}$ and similarly for $P(M|h)$. We will assume for now that $0 \leq \lambda_i, \mu_i \leq 1$.

There are two key ideas of our reduction. In a given context, for a given output, we would like to form an HMM whose probability of that output is proportional to the product of the $\lambda$s of the true indicator functions. That is, we would like the output probability for L to be proportional to $\lambda_1 \times \lambda_2 \times \lambda_3$. By simply creating a chain of states whose transition probabilities are $\lambda_1$, $\lambda_2$, $\lambda_3$, we can achieve this goal.

In Figure 1, we show a simple network that has this first desired behavior, namely that the probability of getting from the beginning to the end is proportional to $\lambda_1 \times \lambda_2 \times \lambda_3$. This HMM is intentionally missing many arcs - we will complete them in a moment. Unless we mention otherwise, all of the transitions we draw are non-emitting transitions, a standard part of the HMM formalism, although not often seen in the machine learning community. (Non-emitting transitions forming loops mean that even with a finite input, we may need to consider infinitely many paths through the HMM.)

Now, the more complicated issue is normalization: we want the sum of the output probabilities to be 1. This is actually easy to accomplish. Given our example, we want the final probability of outputting L to be $\frac{\lambda_1\times\lambda_2\times\lambda_3}{\lambda_1\times\lambda_2\times\lambda_3+\mu_1\times\mu_2\times\mu_3}$ and similarly for outputting M. To construct an HMM with this distribution, we simply use the HMM of Figure 2.

With a bit of thought, one can see that the HMM of Figure 2 produces exactly the desired output distribution. The basic idea is very simple: any path to the end outputting L must have as its last five nodes *start*, $l_1$, $l_2$, $l_3$, *end*. There would be an analogous path to the end, producing M, with the only change being the last five nodes. This means that for every path outputting L, there is a comparable path outputting M, and the ratio of the probability of each of these paths will be in the ratio $\lambda_1 \times \lambda_2 \times \lambda_3$ to $\mu_1 \times \mu_2 \times \mu_3$. Thus, the ratio of the sum of all paths producing L to the sum of all paths producing M must be in the proportion $\lambda_1 \times \lambda_2 \times \lambda_3$ to $\mu_1 \times \mu_2 \times \mu_3$.

Now, for any HMM, the sum of the probabilities of all possible outputs is always 1. Since we know their ratio, and we know their sum, the probabilities must be $\frac{\lambda_1\times\lambda_2\times\lambda_3}{\lambda_1\times\lambda_2\times\lambda_3+\mu_1\times\mu_2\times\mu_3}$ and $\frac{\mu_1\times\mu_2\times\mu_3}{\lambda_1\times\lambda_2\times\lambda_3+\mu_1\times\mu_2\times\mu_3}$, exactly as desired. A slightly more formal proof is given in the appendix of the extended version of the paper.

How do we apply this model for training? In the trivial case where there is only a single element of training data, $x$, $h$, we would simply run the forward-backward algorithm on the network we just created. The training data would be the single token, $x$. To handle multiple pieces of training data, we simply string together several of these networks, as shown in Figure 3. Note that the values of the transition network have to be tied. Just as the same value $\lambda_i$ will be used for computing the probability of many outputs during training, the same transition probability $\lambda_i$ will also be used. This equivalence works fine if all $\lambda_i \leq 1$. If some $\lambda_i > 1$ (and, in typical models, almost all $\lambda_i$ are larger than 1) then we must scale them down. We show how, given a maxent model with indicator functions/coefficients, $f_i$, $\lambda_i$ we can create a new model $\lambda'_i$, $f'_i$ where all $\lambda'_i < 1$.

Call a set of indicator functions such that exactly one function in the set is true for any $x, h$ a *group* (with no relationship to the usual algebraic term "group"). We show that we can scale all of the $\lambda$s in a group by a constant without changing any probabilities.

**Lemma 1** *Given a maximum entropy model* $\lambda_1, ..., \lambda_g$, $f_1, ..., f_g$, *such that for some* $k$, $f_1, ..., f_k$ *form a group, we can create a new model,* $\lambda'_i, f'_i$, *with* $\lambda'_1 = \alpha\lambda_1$, ..., $\lambda'_k = \alpha\lambda_k$, $\lambda'_{k+1} = \lambda_{k+1}$, $\lambda'_{k+2} = \lambda_{k+2}$, ..., $\lambda'_g = \lambda_g$ *and for all* $i$, $f'_i(x|h) = f_i(x|h)$. *Then for all* $x, h$, $P'(x|h) = P(x|h)$.

The proof, using simple algebra, is given in the extended version of the paper. The intuition is simply that since, for a given $h$, exactly one indicator $f$ has been scaled by $\alpha$ for each output $x$, the unnormalized values for each $x$ are all scaled by $\alpha$, and the normalized values are thus the same.

**Lemma 2** *Given a maxent model* $\lambda_1, \lambda_2, ..., \lambda_g$, $f_1$, $f_2, ..., f_g$, *we can create a new model such that every* $f$ *is part of a group and the probabilities are unchanged.*

*Proof* The proof is trivial. The new model is of the form $\lambda'_1 = \lambda_1, ..., \lambda'_g = \lambda_g, \lambda'_{g+1} = 1, \lambda'_{g+2} = 1, ...\lambda'_{2g} = 1, f'_1 = f_1, ..., f'_g = f_g, f'_{g+1} = 1 - f_1, ..., f'_{2g} = 1 - f_g$. We call $f'_{g+1}, ..., f'_{2g}$ *anti-indicators*. Since all of the new $\lambda$s are 1, multiplying by them does not change any probabilities. Clearly, each pair $f'_i, f'_{g+i}$ forms a group. ◇

**Theorem 1** *Given a maximum entropy model* $\lambda_i, f_i$, *we can create a new model* $\lambda'_i, f'_i$ *with the same probability distribution, but with all* $\lambda'_i < 1$.

*Proof* We first apply the grouping lemma, Lemma 2. Then, for a given group $G$, let $\max G$ represent the largest coefficient in the group. We apply the scaling lemma to each group $G$, scaling it by a factor of $\frac{1}{2\max G}$. All coefficients are now less than 1. ◇

Unfortunately, this simple conversion is also very inefficient. It turns out that both our HMM training algorithm (as we will describe later) and the standard maximum entropy training algorithm, Generalized Iterative Scaling (GIS) (Darroch and Ratcliff, 1972), are slowed by a factor, $f^\#$, equal to the largest number of non-zero indicator functions, $\max_{x,h}\sum_i f_i(x,h)$, for any $h$ in the training data. Applying the grouping lemma naively typically significantly increases $f^\#$. But in general, a simple variation on the grouping lemma technique works. For many maxent applications, there are only a few indicator types and at most one indicator of each type is true for a given $x, h$. For instance, in language modeling, there might just be unigram, bigram, and trigram indicators. Thus, a much more practical solution is to put indicator functions into sets such that at most one indicator in the set is true for a given $x, h$. If at most one of $f_1...f_k$ is non-zero, then we can add a new coefficient $\lambda_0 = 1$ and an anti-indicator $f_0(x,h) = 1 - f_1(x|h) - f_2(x|h) - \cdots - f_k(x|h)$, creating a group of $f_0...f_k$ while leaving $f^\#$ and the probability distribution unchanged. When the number of sets such that at most one indicator function is true equals $f^\#$, which is often the case, there is no slowdown to the training algorithms at all from creating the groups and applying the scaling lemma.

We need one more lemma:

**Lemma 3** *Given a model* $\lambda'_i, f'_i$ *to which anti-*

*indicators have been added, we can create a model* $\lambda_i, f_i$ *with the anti-indicators removed, and different* $\lambda$*'s, but the same probability distribution.*

*Proof* For each group $G$, let $\lambda'_G, f'_G$ be the anti-indicator in the group. Rescale the group by $1/\lambda'_G$ (which does not change the probability distribution). Now $\lambda'_G = 1$. Remove $\lambda'_G, f'_G$, which, since the rescaled $\lambda'_G = 1$, does not change the probability distribution. ◇

Note that the space of maxent models is convex with respect to the probability of the training data. This means that we will not encounter local optima or saddle points during our search. Furthermore, there is a unique model in the form of Equation 1 that maximizes the likelihood of the training data and this model will both satisfy the constraints of Equation 2 and will minimize the Kullback-Leibler divergence from the uniform distribution (Della Pietra, Della Pietra, and Lafferty, 1995). The model is unique in the sense of the probabilities it assigns, although there may be many possible settings for the parameter values (as was shown with the scaling lemma). We can now show that for any maxent model, we can train the parameters using an HMM, and get the same model, in terms of probabilities.

**Theorem 2** *Let* $f_i$ *be a set of indicator functions. Train this maxent model to convergence on the training data and call this* $M$*. By adding anti-indicators, create a new set of indicators* $f'_i$ *such that every* $f'_i$ *is part of a group (Lemma 2). Using the same training data, create an HMM of the form shown in Figure 3. Train this HMM to convergence with the forward-backward algorithm (which is guaranteed to converge to a locally optimal point, or to a cycle of equally good locally optimal points, in terms of probabilities). The probabilities of this HMM are the same as the probabilities of the maxent model,* $M$*.*

*Proof* First, we need to consider another maxent model, $M'$, using the indicator functions $f'_i$ with parameters $\lambda'_i$ trained on the training data. For any model with indicator functions $f_i$, there is an equivalent model (in terms of probabilities) for indicator functions $f'_i$ and vice versa. The first direction comes from simply setting the $\lambda'_i$ for each anti-indicator to 1, and the opposite direction comes from Lemma 3. Since $M$ and $M'$ are both global optima, and since the global optimum is unique (in terms of probabilities), $M$ and $M'$ must yield the same probability distribution.

Now, we notice that the space of HMMs in the form of Figure 3 is convex. This follows since, for a given setting of the parameters in this space, there is a maxent model with the same parameters, assigning the same probabilities to all $x, h$. While the space of HMMs is smaller than the space of maxent models with these parameters (since the parameters must all be $\leq 1$), this subspace is still convex. Thus, there is a global minimum in this space. Apply the scaling lemma to $M'$, yielding $M''$, which has the same probability distribution as $M$ and $M'$. $M''$ is in the HMM subspace and at the unique global minimum. The HMM we trained is also at the global minimum within its subsapce. Thus the HMM has the same probability distribution as $M''$. ◇

One might fear that training the HMMs we describe would be a complex or slow task, since solving the equations for the forward-backward algorithm when the network has non-emitting transitions can require inversion of a matrix. For all of the models described here, except for Maximum Entropy Markov Models, the equations can be solved with no matrix inversions. The equations are easy to derive: one simply writes out the infinite sums, which are geometric series that can be reduced to closed forms with simple algebra. However, because of the number of cases, they are a bit too complex to include here. Still, they require only a reasonable number of operations, on the order of 50 per training example, for a two-output model. The full equations are derived in the extended version of this paper.

One might hope that the HMM training algorithms would be more efficient than the maxent ones. Maxent models are typically trained using the GIS algorithm (Darroch and Ratcliff, 1972) or variations thereon (Lafferty and Suhm, 1995; Goodman, 2001; Della Pietra, Della Pietra, and Lafferty, 1995). The basic idea is to simply update each $\lambda_i$ by the ratio between the observed frequency of $f_i$ and the expected frequency. However, to guarantee convergence, the step size is slowed by a factor proportional to $f^\#$, which is the maximum number of indicator functions that can be true for any $x$,$h$. On the other hand, the forward-backward algorithm does not contain any such factor, so its speed appears at first glance independent of the number of active constraints. Unfortunately, it turns out that the update speed for transitions, other than the ones in the $\lambda_3$ or $\mu_3$ position (the final position), can be very slow. In practice, one can rotate through the mapping of parameters to HMM topologies, so that each constraint type ends up in the final position. If there are $f^\#$ constraint classes, then after $f^\#$ iterations, each class will have been in the final position once, meaning the learning is slowed by a factor of $f^\#$, the same factor as for GIS. (Note that Improved Iterative Scaling (Della Pietra, Della Pietra, and Lafferty, 1997) has a similar, though sometimes smaller slow down. In particular, for IIS, there is a

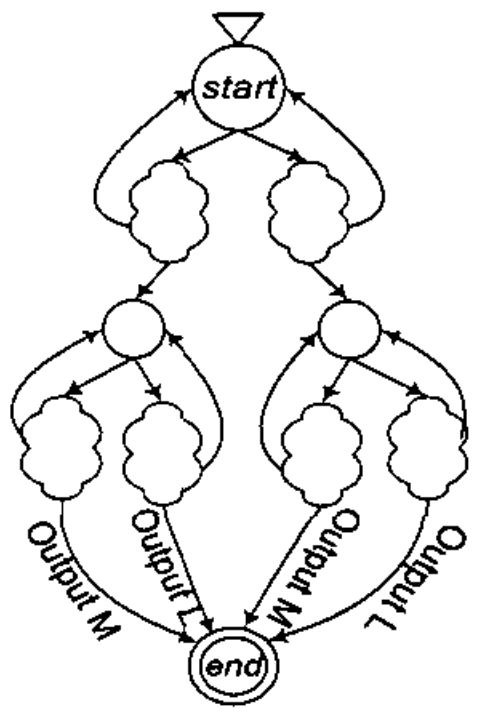

Figure 4: Hidden Variable Maxent Model

term equivalent to $f^{\#}$, but on a per-training example basis, rather than globally.)

## 3 Extensions

Maxent models are powerful models that can combine a large variety of information. Unfortunately, they are limited in certain important ways. In particular, they cannot capture hidden variables. We can extend the maxent formalism in such a way that hidden variables can be captured, and these extensions are also easy to model using HMMs, leading immediately to training algorithms with guaranteed local optimality (or saddle points). First, we show how to make maxent models with a simple hidden variable. Then, we show how to make HMMs where the transition probability depends on a maxent model. Next, we quickly describe a recently introduced model, Maximum Entropy Markov Models (MEMMs), which are essentially maxent models with hidden state sequences. MEMMs too can be expressed as HMMs. We then show that a variation on MEMMs, Conditional Random Fields (CRFs), also can be reduced to HMMs. Finally, we show that certain types of continuous variables also fit into this framework. In the extended paper, we also show that even Probabilistic Context Free Grammars can be combined with maximum entropy models to form powerful new models that further extend the power of the framework.

We consider a simple example, a hidden variable that takes a value, N or P, and then produces an output, L or M, the probability of which depends on the value of the hidden variable, and the conditioning context. Both the probability of the value of the hidden variable, and the output given the hidden variable, depend on maximum entropy models. We can build a network such as is shown in Figure 4. Each "cloud" in this network represents one half of the network of Figure 2. In this network, the first pair of clouds selects the value of the hidden variable, either N or P. In the left subnetwork, used when the value of the hidden variable is N, the indicator functions that apply are those that are true for the hidden variable being N, while on the right, they are the indicators when the hidden variable is P.

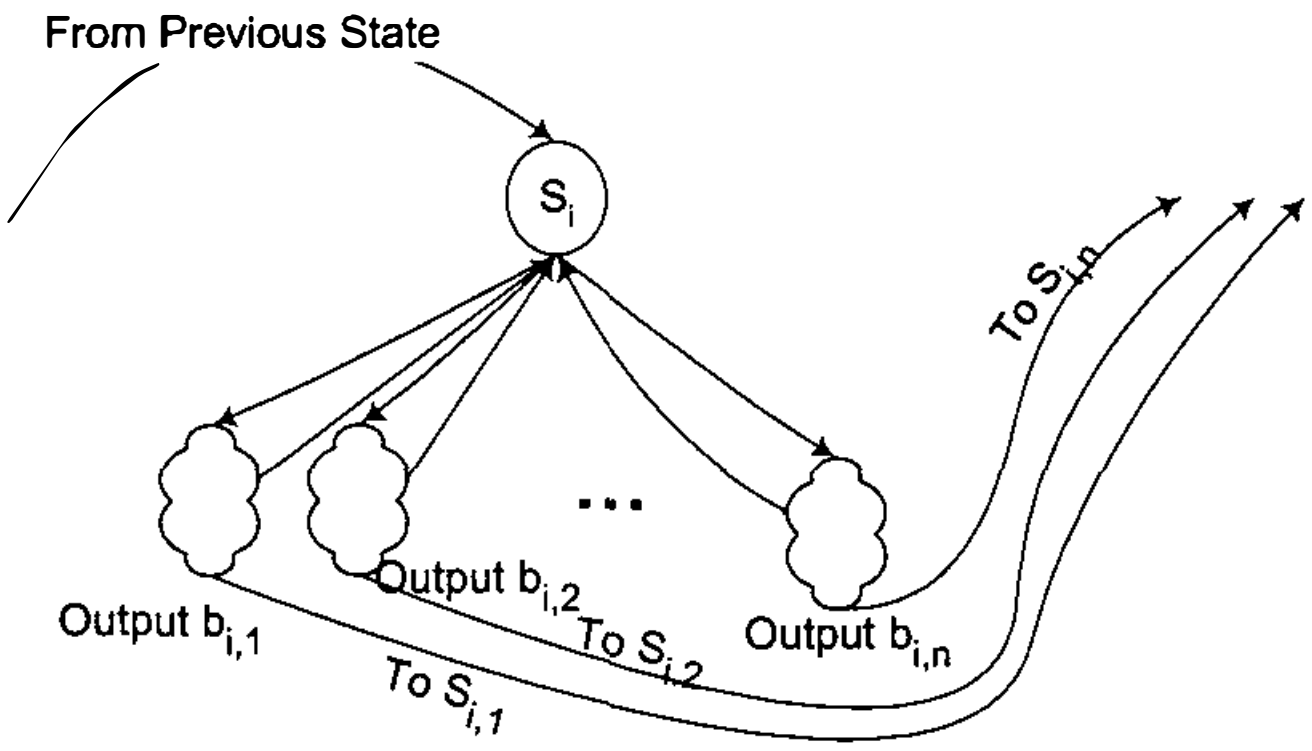

Figure 5: Hidden Markov Model with Maxent Transition/Outputs

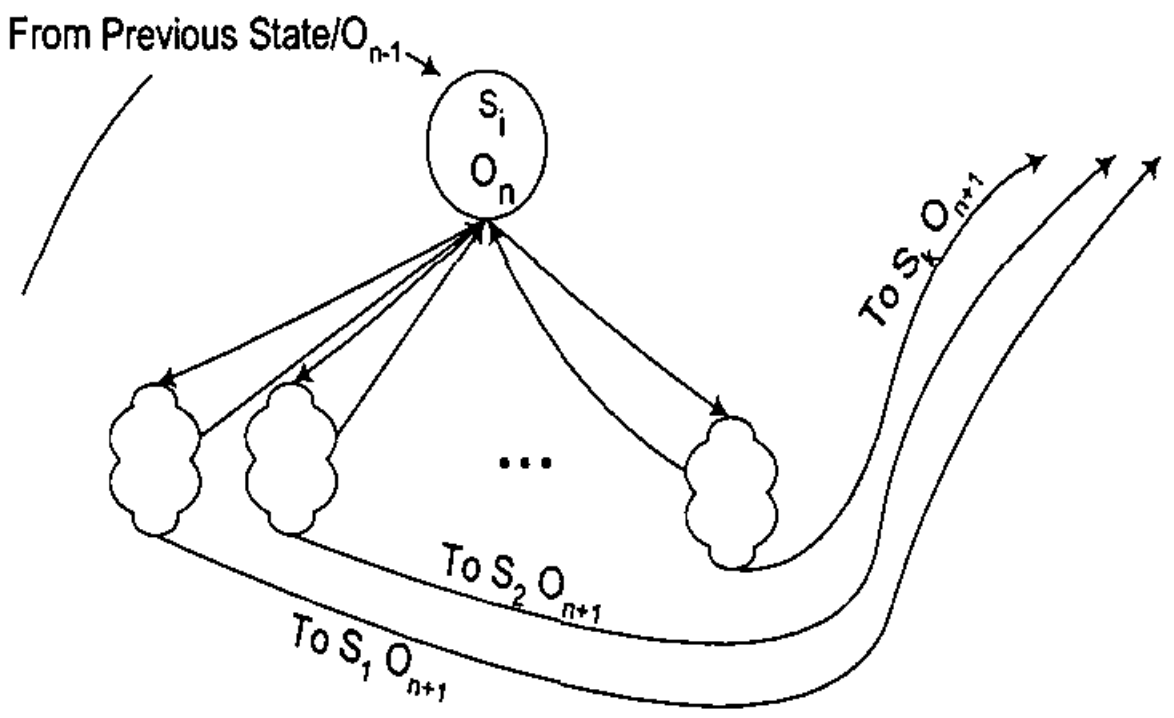

Figure 6: Subnetwork for Maximum Entropy Markov Model

Our equivalence starts to become much more interesting when we consider hidden state sequences. We now go on to describe briefly how to build an HMM where the transition probabilities and output probabilities depend on a maxent model, as shown in Figure 5. All transitions out of the inside of a cloud, of which there may be many, go back to the beginning of the state. Eventually, the model reaches the end of a cloud, produces an output $b_{i,j}$, and transitions to the beginning of the next state $S_{i,j}$. The size of the constructed HMM is proportional to $f^{\#}$ (which determines the size of each cloud) times the number of transitions (which determines the number of clouds.)

Recently, another formalism combining HMMs with maxent models, Maximum Entropy Markov Models

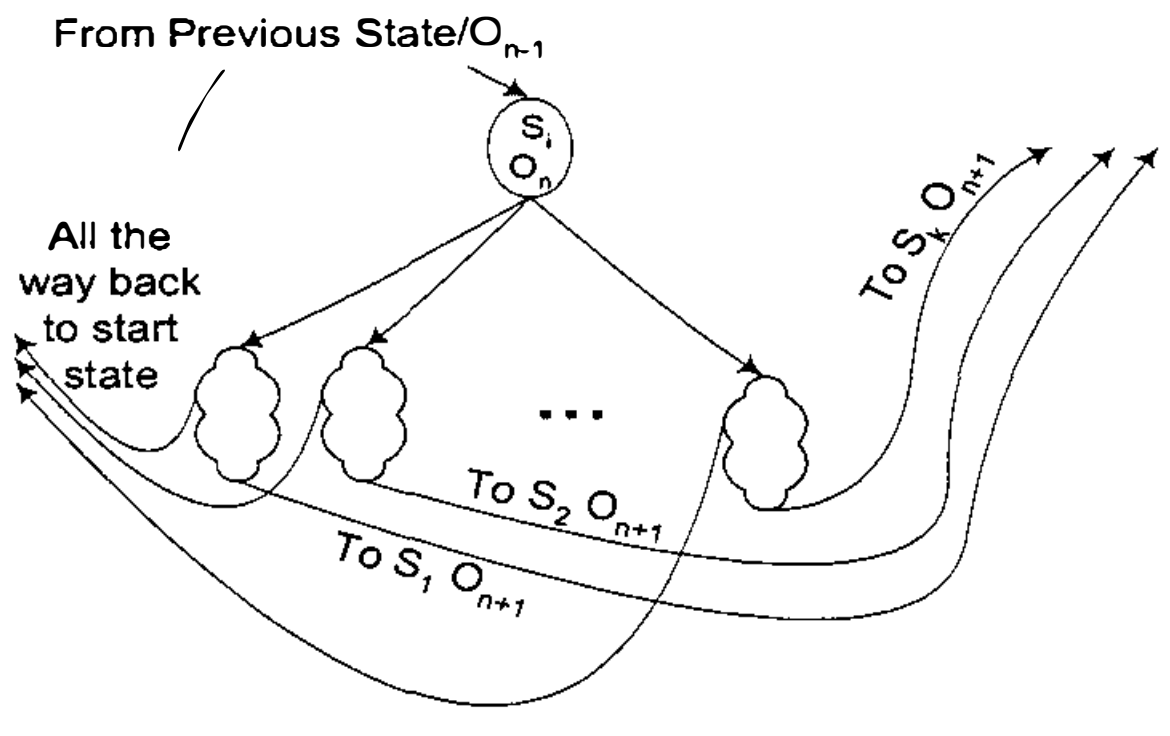

Figure 7: Subnetwork for Conditional Random Field Model

(MEMMs) has been shown to outperform HMMs for a text segmentation task (McCallum, Freitag, and Pereira, 2000). MEMMs are similar to the model we just described, except that the state transitions are conditioned on the observed outputs, rather than producing outputs. Thus, this model assigns probabilities to state sequences, conditioned on observation sequences, rather than conventional joint probabilities.

MEMMs can be easily described using our formalism, as shown in Figure 6, which corresponds to the transitions for a single state, $S_i$ at a given observation $O_n$. Because we build this model on the fly, after the outputs are already known, the model itself does not produce any outputs, but there are different sub-networks for each state, observed output pair. The clouds represent HMM states and transitions corresponding to the appropriate indicator functions for the $S_i$, $O_j$, and the corresponding next state. Any transitions that would normally return to the start state return to the entrance for this sub-network, ensuring that the probabilities are correctly normalized.

Even more recently, a variation on MEMMs has been proposed, namely Conditional Random Fields (Lafferty, McCallum et al. 2001); this model does not normalize the output of each state, but instead normalizes based on all possible state sequences. In Figure 7, we show that this can be considered a simple variation on MEMMs. The only difference is that, since we wish to normalize at the state sequence level, instead of within states, any return arcs of the sub-networks return all the way to the start state at the beginning of the sequence, rather than to an internal state. Note that training this model using the forward-backward algorithm is probably impractical.[2]

[2]As we will discuss in Section 4, efficient training using the forward-backward algorithm probably requires reordering states on different iterations; it is not clear how to do this reordering for Conditional Random Fields.

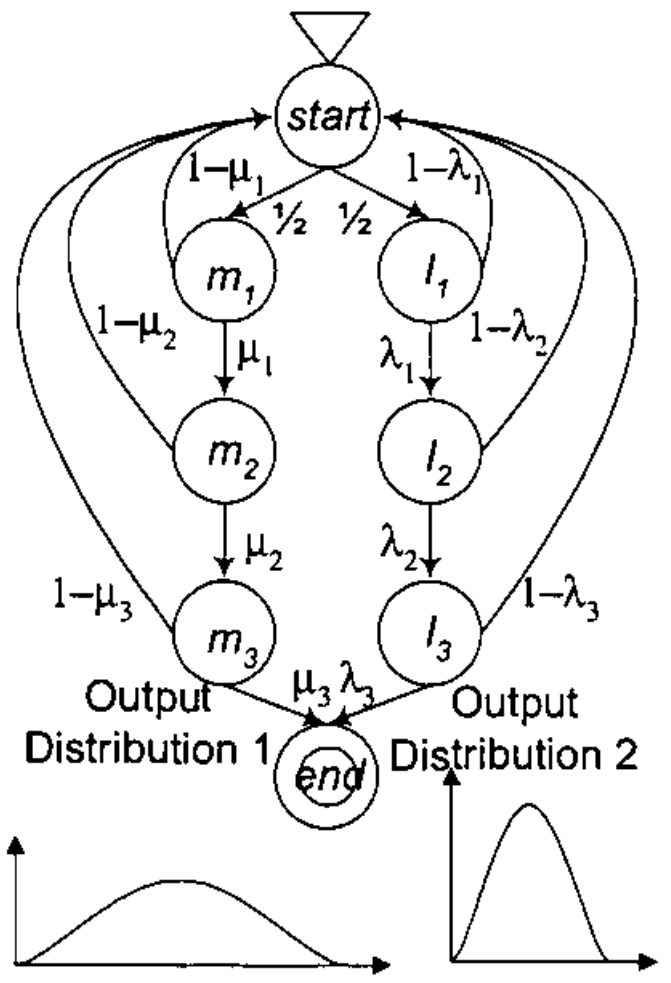

Figure 8: Continuous Outputs

The maximum entropy framework is usually used for discrete outputs. However, just as HMMs can be used to produce continuous outputs, we can extend the maxent framework to also produce continuous outputs. The continuous distribution will be a mixture model, with mixture weights that depend on a maxent model. We simply create an HMM that produces a continuous output, e.g. a Gaussian distribution, instead of a discrete output, as shown in Figure 8. The output can be one-dimensional, or multi-dimensional. Such a model could be useful in a variety of fields. For instance, in speech recognition, the output could correspond to the acoustic output of the HMM; the maxent model could replace the phonetic decision tree typically used to determine which gaussians are output. The constraints of the model could include the gender of the speaker, channel characteristics, phonetic questions such as those used in phonetic decision trees, and any other measurable data, such as speaking rate or pitch, if these can be accurately estimated. In the extended paper, we show how to create maxent HMMs with hidden state sequences. Similarly, we can integrate this model into a larger HMM with hidden state sequences and continous outputs, allowing the entire speech recognition system to be modeled as a single large HMM, and also allowing reestimation of all parameters simultaneously, leading to a locally optimal model; the decision trees normally used have no optimality guarantees.

## 4 Experimental Results

One of the most important contributions of this paper is to make it easy to train models that extend the maximum entropy framework. In this section, we very briefly describe some experimental results using one of these extensions. More details are given in the extended version of the paper. We have been exploring automated grammar checking using machine learning. Standard machine learning algorithms do well on many grammar-checking problems ("their/they're/there" or "than/then"), but have trouble with tasks like subject-verb agreement, which requires identifying the subject and then determining if there is agreement. Data annotated with sentence subjects is unavailable in most languages and expensive to create, so we treat the subject as a hidden variable, whose value we must learn. We also need to know if the subject is singular or plural. We consider this information to also be hidden. Thus, we have a complex hidden variable, the value of which determines the subject of the sentence and its number (singular/plural). We build models of a form similar to that in Figure 4, a maximum entropy model with a hidden variable. The actual model we need is slightly simpler, since, given the value of the number component of this hidden variable, it is trivial to determine the correct output: singular or plural, meaning we do not need the "clouds" on the bottom of this model.

| *technique* | *standard set* | *best extended set* |
|---|---|---|
| hidden variable maxent | | **89.1%** |
| transformation-based | 87.4% | 88.3% |
| decision list | 87.0% | 87.8% |

Figure 9: Subject-Verb Results

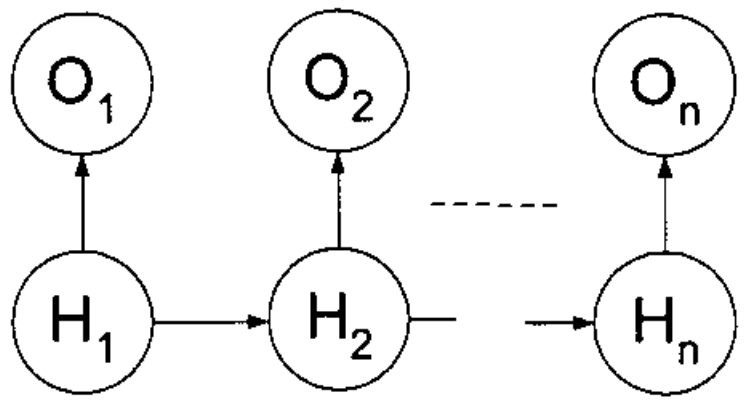

Figure 10: Graphical Model for an HMM (no non-emitting states)

We considered subjects in a window of $\pm 9$ words. For feature functions $f_i$, we used the position of the word, the part-of-speech tag of the word, the part-of-speech tag of its four closest neighbors, and the identity of the word itself. We considered only the verbs "is" and "are" for simplicity.

We compared this to a variety of models trained using both transformation-based learning and probabilistic decision lists, both of which have been shown to work well for word-disambiguation tasks of this sort in the past (Banko and Brill, 2001). We used both standard feature types, previously used with these models, and more complex feature sets hand-optimized for this task. The results are shown in Figure 9.

## 5 Discussion

It is interesting to compare the HMM framework to the Bayesian Network framework. There does not appear to be any representation of maxent models in the Bayes Net framework analogous to the one here. While graphical models can show the dependencies between inputs and outputs in an abstract sense, more formal Bayesian Network models with parameters that can be reestimated by EM do not allow for the looping representation critical to our transformation. In particular, recall that Bayesian Networks can represent HMMs without non-emitting arcs by using one variable to represent the state for each time, as shown in Figure 10. However, with non-emitting arcs, an unlimited number of transitions can occur at each time, which would require an infinite number of states to represent as a Bayesian Network, as shown in Figure 11. It is possible to convert an HMM with non-emitting arcs to one without non-emitting arcs, but the conversion process destroys the tieing between the HMM and maxent parameters. There are extensions to the Bayes Net framework (Koller, McAllester, and Pfeffer, 1997) that do allow looping. Unfortunately, so far, there are not closed form solutions for these extensions. In the extended version of this paper, we discuss all of these, and other possible ways to represent maxent models with graphical models in more detail, concluding that there is no comparably simple reduction for graphical models. This implies that for a variety of interesting model types, it is advantageous to think in terms of HMMs instead of in terms of Bayes Nets.

This paper, by showing the relationship between HMMs and maxent models, opens up a variety of new questions. Can important techniques from maxent models, such as recent smoothing techniques (Chen and Rosenfeld, 1999), be extended to the HMM framework? If so, would they be useful for HMMs in general? Are there HMM smoothing or training techniques that could be applied to smoothing these models?

Maxent models have been in use in statistical NLP for many years now. Recent improvements, such as new smoothing techniques (Chen and Rosenfeld, 1999), and recent speedup techniques (Lafferty and Suhm, 1995; Goodman, 2001), are finally making interesting maxent models possible. Perhaps the techniques of this paper, allowing maxent models to be used for more problems, with a variety of hidden variables, and with more complex models will lead to important practical gains. We have already shown that techniques inspired by this equivalence outperform standard techniques on an interesting problem, subject-verb agreement.

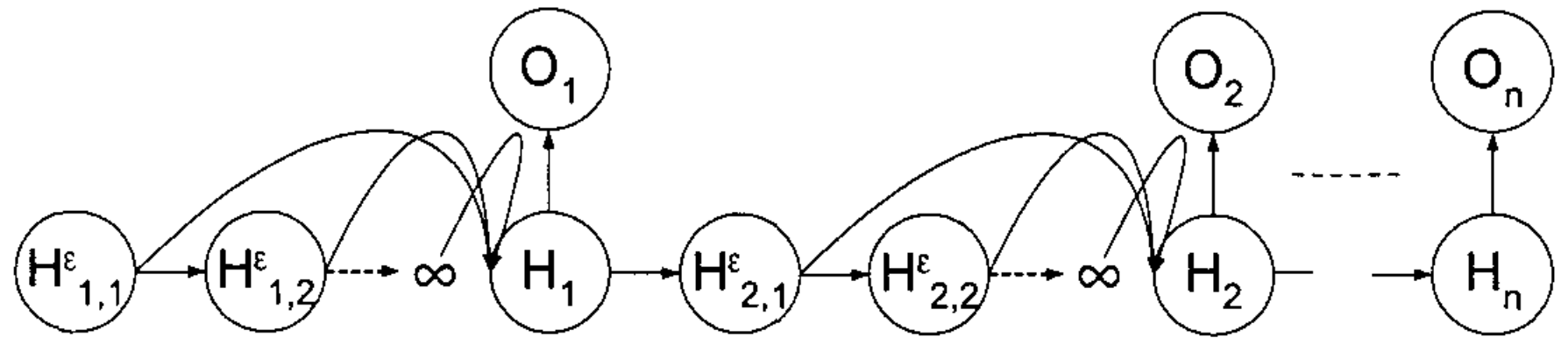

Figure 11: Graphical Model for an HMM with non-emitting states

Several reductions related to this one are well known and widely useful from a theoretical standpoint, such as the reducibility of HMMs (without non-emitting arcs) to Bayesian Networks, or of Probabilistic Context Free Grammars to HMMs with stacks. We hope that the reduction described here will be at least as useful for understanding the relationship between HMMs and the models we describe. In particular, this paper shows that a large class of interesting models previously thought to be distinct can be unified as part of a well known, well understood formalism, HMMs. These models include standard maxent models, MEMMs, CRFs, maxent models with hidden variables, maxent models with continuous outputs, and maxent models with hidden state sequences. Phrasing these models as HMMs leads immediately to training algorithms, with guaranteed convergence properties, but more importantly, leads to a better understanding of the models themselves, and of their relationships.

**Acknowledgments**

We wish to thank many people for their advice and encouragement, including David Heckerman, Chris Meek, Oliver Downs, Sam Roweis, Fernando Pereira, Harriet Nock, and Ciprian Chelba. We would like to thank Stan Chen for his advice and discouragement.